\documentclass[letterpaper, 10 pt, conference]{ieeeconf}  

\IEEEoverridecommandlockouts                              

\usepackage{amsmath} 
\usepackage{amssymb}  
\usepackage{kotex}  
\usepackage{graphicx}
\usepackage{listings}
\usepackage{multirow} 
\usepackage{booktabs}
\usepackage{colortbl}
\usepackage{xcolor}
\usepackage[normalem]{ulem}
\usepackage{caption}
\let\labelindent\relax
\usepackage{enumitem}
\setlist[itemize]
{leftmargin=1.0em}

\usepackage[absolute,overlay]{textpos}
\usepackage{cite}

\lstdefinelanguage{json}{
    basicstyle=\ttfamily,
    numbers=left,
    numberstyle=\tiny,
    stepnumber=1,
    numbersep=5pt,
    showstringspaces=false,
    breaklines=true,
    frame=lines,
    literate=
     *{true}{{{\color{blue}true}}}{4}
      {false}{{{\color{blue}false}}}{5}
      {null}{{{\color{blue}null}}}{4}
      {:}{{{\color{red}{:}}}}{1}
      {,}{{{\color{red}{,}}}}{1}
}

\title{\LARGE \bf
ESPADA: Execution Speedup via Semantics Aware Demonstration Data Downsampling for Imitation Learning
}
\author{Byungju Kim$^{1,2,*}$, Jinu Pahk$^{1,2,*}$, Chungwoo Lee$^{1,*}$, Jaejoon Kim$^{1,2,*}$, Jangha Lee$^{1,2,*}$, Theo Taeyeong Kim$^{1,2}$, \\ Kyuhwan Shim$^{2}$, Jun Ki Lee$^{2}$, Byoung-Tak Zhang$^{1,2,\dagger}$
\thanks{$^{1}$Tommoro Robotics}
\thanks{$^{2}$Seoul National University}
\thanks{* equal contribution. $\dagger$ corresponding authors.}
\thanks{This work was supported by the Technology Innovation Program (RS-2025-25453780, Development of a National Humanoid AI Robot Foundation Model for Multi-Task Applications) funded by the Ministry of Trade, Industry and Resources (MOTIE, Korea).}
}

\begin{document}

\maketitle
\thispagestyle{empty}
\pagestyle{empty}


\begin{abstract}

Behavior-cloning based visuomotor policies enable precise manipulation but often inherit the slow, cautious tempo of human demonstrations, limiting practical deployment.
However, prior studies on acceleration methods mainly rely on statistical or heuristic cues that ignore task semantics and can fail across diverse manipulation settings.
We present \textbf{ESPADA}, a semantic and spatially aware framework that segments demonstrations using a VLM--LLM pipeline with 3D gripper–object relations, enabling aggressive downsampling only in non-critical segments while preserving precision-critical phases, without requiring extra data or architectural modifications, or any form of retraining.
To scale from a single annotated episode to the full dataset, ESPADA propagates segment labels via Dynamic Time Warping (DTW) on dynamics-only features.
Across both simulation and real-world experiments with ACT and DP baselines, ESPADA achieves approximately a 2x speed-up while maintaining success rates, narrowing the gap between human demonstrations and efficient robot control.

\end{abstract}

\section{INTRODUCTION}

Imitation learning (IL) has become a central paradigm in robot learning~\cite{act, dp, pi0, alohaunleashed, 3ddp, robomimic, bridge}, enabling robots to acquire manipulation skills from expert demonstrations without explicit reward design or costly exploration. While early applications focused on simple pick-and-place tasks, recent advances have extended IL to long-horizon~\cite{pi0}, contact-rich~\cite{factr}, and visually complex manipulation~\cite{act, openx}. Policies such as Action Chunking Transformer (ACT)~\cite{act} and Diffusion Policy (DP)~\cite{dp} demonstrate the practicality of this paradigm and serve as strong baselines for visuomotor manipulation.

Despite these successes, IL policies often execute tasks slowly. Human demonstrators typically act cautiously to ensure safety and maximize success rates, and prior work has intentionally adopted slow demonstrations due to (i) camera frame-rate constraints, (ii) improved training stability under slower motions, and (iii) the anthropomorphism gap between human and robot kinematics~\cite{wildrobot}. Consequently, demonstrations become temporally saturated, causing learned policies to inherit unnecessarily slow execution speeds~\cite{subconscious}.

\begin{figure}[t]
    \centering
    \includegraphics[width=0.8\columnwidth]{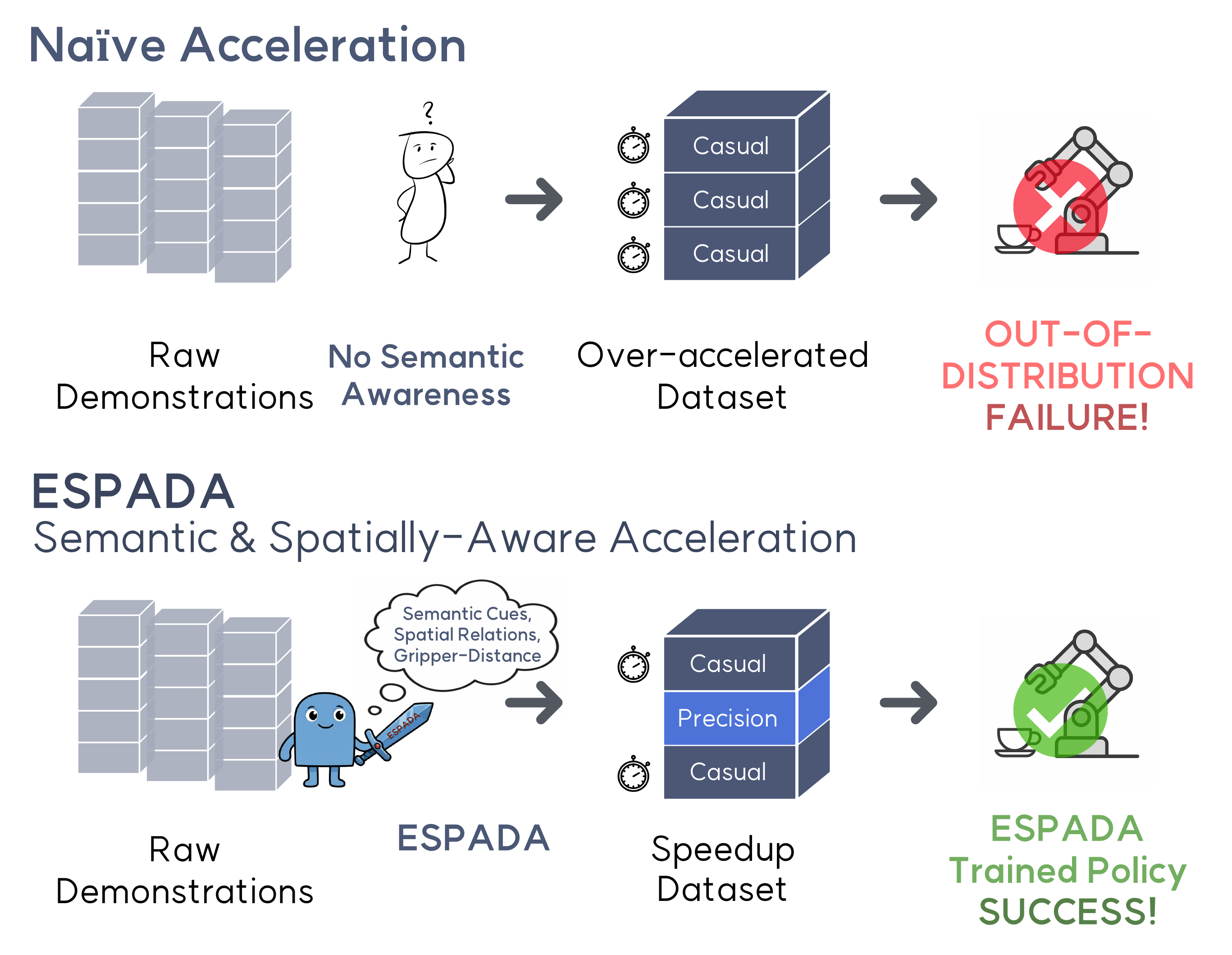}
    \caption{\textbf{Naïve and heuristic-based acceleration breaks precision behavior in manipulation tasks.} ESPADA uses semantics and 3D spatial cues to preserve contact-critical phases while accelerating transit motions.}
    \label{fig:espada-vs-dsu}
\end{figure}

Naïvely accelerating demonstrations—for example by replaying trajectories faster or uniformly subsampling observations—often pushes policies out of distribution and degrades performance. To address this, several methods attempt to identify segments that can be safely accelerated. For example, SAIL~\cite{sail} uses clustering over AWE features~\cite{awe} to discover coarse phases, while DemoSpeedup~\cite{demospeedup} estimates action entropy using a proxy policy and treats high-entropy segments as accelerable. However, these approaches rely on heuristic assumptions about motion statistics rather than task semantics, making them fragile when scene conditions change or demonstrations become multimodal.
        
In particular, action entropy is not a reliable proxy for manipulation precision. Multimodal strategies arising from scene variability may produce high entropy despite strict precision requirements, while repetitive motions may exhibit low entropy even when acceleration is safe. As a result, entropy-based segmentation can misidentify precision-critical phases.



Accordingly, we introduce \textbf{ESPADA} \textit{(Execution Speedup
via Spatially Aware Demonstration Data Downsampling for
Imitation Learning)}, a semantic-driven trajectory segmentation framework that 
selectively accelerates demonstrations without extra hardware, additional data, or additional policy 
trainings. Prior methods rely on heuristic motion statistics, such as density clusters or action entropy, 
to implicitly approximate precision. In contrast, ESPADA replaces these assumptions with explicit scene 
semantics and gripper–object 3D relations. These cues reveal the task intent (e.g., approach, align, 
adjust) and the actual interaction state between the gripper and the target object, enabling the system 
to determine exactly where acceleration is safe while preserving genuine precision-critical phases.

Concretely, we extract per-frame 3D coordinates of grippers and key objects using open-vocabulary segmentation~\cite{groundingdino,sam2} and video-based depth estimation~\cite{vda,depthanythingv2}, leveraging temporal context for stable geometry while remaining compatible with monocular visuomotor setups.
These cues, combined with image observations, are summarized by a vision–language model~\cite{internvl3.5} and converted into a compact language representation for LLM-based segmentation. The LLM then classifies segments into \emph{casual} and \emph{precision} based on semantic descriptions and gripper–object distance trends.
Finally, we accelerate casual segments via \emph{replicate-before-downsample} (RBD) with geometric consistency~\cite{demospeedup}, enabling efficient compression with a reproducible, relation-driven labeling policy.

Our contributions are three-fold: 
(i) The first semantic and 3D-relation–aware policy acceleration framework via demonstration downsampling without any additional sensor data or retraining.
(ii) A scalable label transfer scheme that propagates segment labels from a single annotated episode to the rest of the dataset using banded DTW.
(iii) Experimental validation in simulation and real-world settings, with up to a 3.6× execution speedup while maintaining or improving success rates.

\section{Related Work}
\paragraph*{Speeding up imitation learning execution}
Modern visuomotor policies such as ACT~\cite{act} and DP~\cite{dp} achieve strong manipulation performance but typically inherit the slow timing of human demonstrations. Naïve acceleration through uniform subsampling or increased control rates can push policies out of distribution and degrade performance. Recent approaches therefore attempt to identify segments that can be safely accelerated. DemoSpeedup estimates action entropy using a proxy policy and accelerates high-entropy segments with geometric constraints~\cite{demospeedup}, while SAIL segments trajectories using waypoint complexity and clustering~\cite{sail}. However, these methods rely on heuristic proxies such as entropy or clustering thresholds, which can be fragile under changing scenes or multimodal demonstrations. Our approach instead performs segmentation based on semantic reasoning over spatial relations, producing segments that better align with manipulation intent.

\paragraph*{Temporal segmentation and phase discovery}
Classical phase-discovery pipelines often rely on fixed features and clustering (e.g., DBSCAN~\cite{dbscan}) to recover phases from kinematics or vision (e.g., RoboSubtaskNet~\cite{robosubtasknet}), sometimes assisted by motion primitives. These pipelines can be brittle across tasks and cameras because phase boundaries depend on feature scaling and neighborhood thresholds. Other approaches use latent structure learning for phase discovery (e.g., Temporal Alignment for Control~\cite{taco}), but they still struggle to distinguish precision-critical contact phases from benign transits. ESPADA instead uses 3D gripper--object distance trends as grounded signals, deferring semantic interpretation to a large language model (LLM), which produces coherent manipulation chunks.
\paragraph*{Vision--language for robotics}
Vision–language models combine grounded perception with structured reasoning and have been explored in generalist robotic systems~\cite{rt1, openx, saycan, palm}. We adopt a modular VLM$\rightarrow$LLM pipeline consisting of open-vocabulary segmentation (Grounded DINO + SAM~\cite{groundingdino,sam}), monocular depth estimation~\cite{vda, depthanythingv2}, and semantic scene summaries (InternVL~\cite{internvl3.5}). The extracted spatial relations are converted into language tokens so that an LLM can reason over gripper–object geometry and temporal trends. This modular design improves interpretability and benefits directly from advances in foundation models.

\paragraph*{Positioning}
ESPADA addresses key limitations of entropy- or clustering-based acceleration methods such as DemoSpeedup and SAIL. Rather than assuming that entropy or motion complexity reliably indicates casual motion, ESPADA relies on explicit relational cues—particularly gripper–object distance trends and scene semantics—to determine when acceleration is safe. This produces more coherent temporal segments and reduces sensitivity to clustering hyperparameters, leading to more stable compression decisions.



\begin{figure*}[t]
    \centering
    \includegraphics[width=\textwidth]{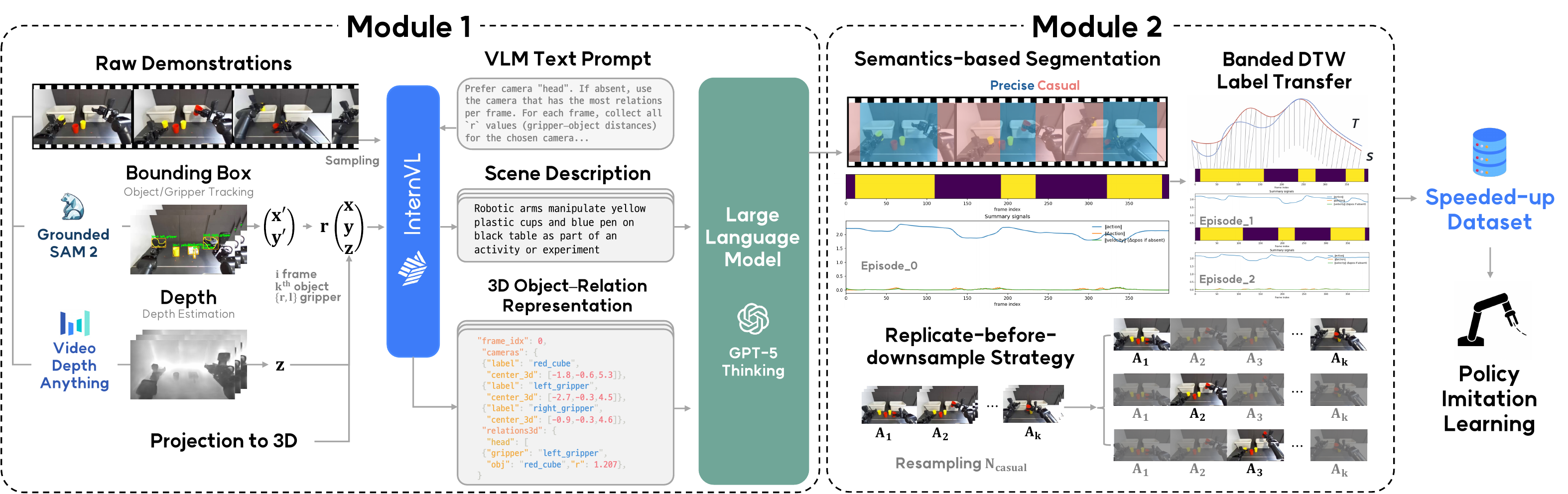}
    \caption{
        \textbf{Overview of ESPADA}. 
        We use Grounded-SAM2 and Video Depth Anything (VDA) to extract 3D object-gripper relations, 
        summarize the episode with a VLM, and segment trajectories with an LLM into \texttt{precision} 
        and \texttt{casual} spans. Segment-wise downsampling is then applied with replicate-before-downsample 
        and geometric consistency, producing faster yet safe demonstrations for imitation learning. 
        To reduce annotation cost, we annotate only episode~0 via the VLM$\rightarrow$LLM pipeline, and 
        propagate its labels to other episodes with \emph{banded DTW label transfer}, which aligns action 
        sequences under temporal variation while refining boundaries.
    }
    \label{fig:main}
\end{figure*}


\section{Problem Setup}

We consider a dataset of robot manipulation demonstrations
\(
\mathcal{D} = \{ (o_t, a_t) \}_{t=1}^{T}
\),
where $o_t$ are observations (RGB images, proprioception) 
and $a_t$ are low-level actions (joint position commands). 
Demonstrations are collected at control frequencies 
$f_{\mathrm{ctrl}} \in [30,50]$ Hz, producing temporally dense trajectories. 
Policies such as ACT~\cite{act} and DP~\cite{dp} predict fixed-horizon action chunks 
$A_t = \{a_{t},\ldots,a_{t+K-1}\}$ from recent observations.

A core issue is that human demonstrations are performed slowly and cautiously, 
yielding oversampled sequences. 
Uniformly downsampling often pushes trajectories out-of-distribution, because aggressive temporal thinning alters the local action–state transitions seen during training, introduces temporal aliasing in contact-rich or high-curvature segments, and disrupts the smoothness assumptions under which behavior-cloned policies generalize.
Our goal is to accelerate demonstrations offline by selectively reducing temporal density in \textit{casual phases} while applying only mild reduction in \textit{precision-critical phases}, 
without modifying the runtime control loop or the policy architecture.

Formally, we segment each trajectory as
\(
\mathcal{S} = \{(s_i,e_i,y_i)\}_{i=1}^M
\),
with $y_i \in \{\texttt{casual},  \texttt{precision}\}$. 
We then transform $\mathcal{D}$ by
\[
\mathcal{T}(\mathcal{D},\mathcal{S},N) =
\bigcup_{i=1}^M
\begin{cases}
\text{RBD}(\mathcal{D}[s_i:e_i],N_{y_i})
\end{cases}
\]
where \(y_i \in \{\texttt{precision, casual}\}\), and RBD ensures that all original frames 
are preserved across replicas during downsampling. Here, \texttt{casual} indicates segments that can be safely downsampled without compromising task fidelity, while \texttt{precision} denotes 
precision-critical spans that are retained at near full resolution, with only minimal acceleration applied when safe. For stability, we enforce \emph{geometric consistency} \cite{demospeedup} by adjusting accelerated chunk horizons 
$K'$ so that the spatial displacement 
$\sum_{k=0}^{K'-1}\|\Delta \mathbf{x}_{t+k}\|$ 
matches that of the original horizon $K$. \(N_{y_i}\) denotes the number of replicas in RBD, determined by the maximum acceleration ratio.


\begin{figure*}[t]
    \centering
    \includegraphics[width=0.8\textwidth]{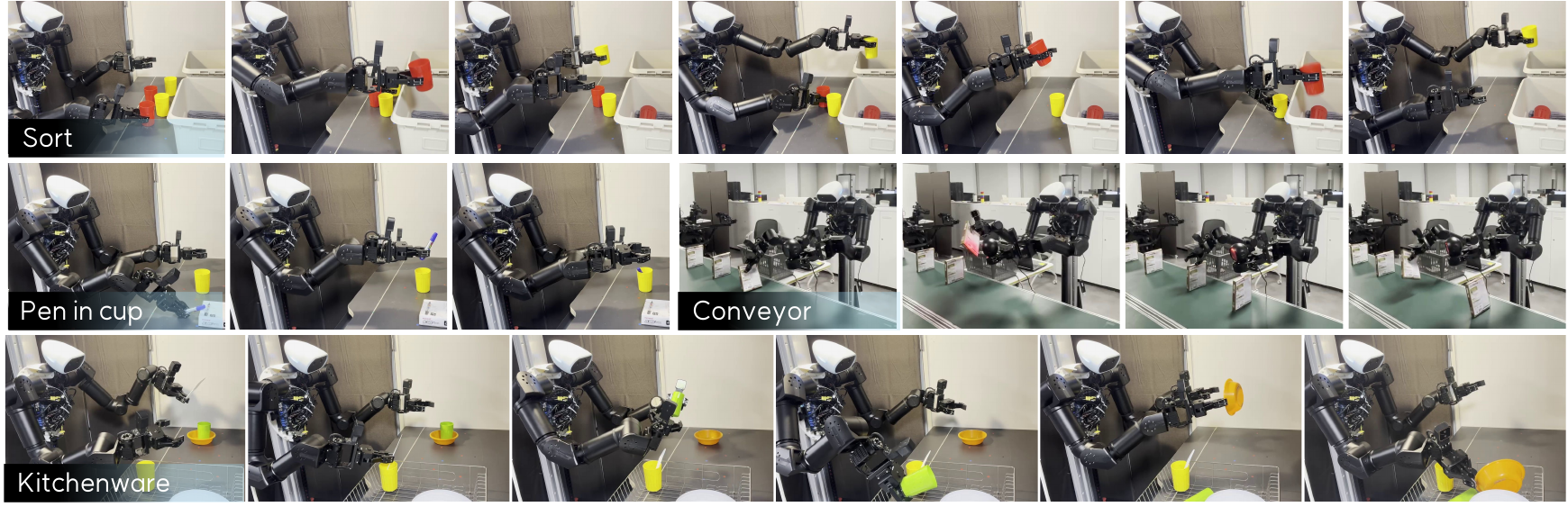}
    \caption{\textbf{Real-world evaluation of ESPADA on the AI Worker robot across four representative manipulation tasks}. (i) Sort – classifying colored objects into bins, (ii) Pen in cup – placing a pen into a cup, (iii) Conveyor – transferring curry into a basket along a moving belt, and (iv) Kitchenware – handling bowls and cups.}
    \label{fig:realworld-exp}
\end{figure*}

\section{Method}
Our pipeline converts raw demonstrations into \emph{semantically and spatially informed} segments that can be selectively accelerated, then constructs an acceleration-aware training set via replicate-before-downsample (RBD) with geometric consistency. Figure~\ref{fig:main} provides an overview.
\subsection{Context- and Spatial-Aware Segmentation via VLM $\rightarrow$ LLM}

\paragraph{Object tracking with interactive keyframe seeding.}
We obtain open-vocabulary object tracks from demonstration videos using Grounded-SAM2~\cite{sam2,groundingdino}. Users may optionally provide sparse keyframe annotations (boxes or point groups) via a lightweight UI. A label–ID mapping is maintained across keyframes, and IoU-based association propagates user labels to SAM2 track IDs. To maintain track continuity, we use a keep-alive strategy for short outages and periodically re-run Grounding DINO to recover lost tracks.

To bootstrap object grounding, we sample $\sim$10 frames from episode 0 and use InternVL~\cite{internvl3.5} to generate a compact task description. Grounding DINO then detects task-relevant entities (e.g., \texttt{left\_gripper}, \texttt{right\_gripper}, target objects), whose boxes initialize SAM2 for mask and bounding-box propagation across the episode.

If tracking fails due to occlusion or the object leaving the camera view, we optionally allow a lightweight manual correction (a single bounding box prompt) to reinitialize the tracker. In practice, automatic tracking already succeeds for the majority of frames (about 89\% on average), so manual intervention is rarely required.

\paragraph{Depth estimation and 3D back-projection.}
We estimate per-frame depths with VDA/DA2~\cite{vda,depthanythingv2} (metric or relative; optionally scaled by a factor $z_{\mathrm{scale}}$). As we obtained the pixel coordinates (u, v) of each object of interest in the previous step, given the corresponding depth Z, we can recover its 3D position in the camera coordinate frame via standard back-projection:
\begin{equation}
\mathbf{p} = Z K^{-1}[u, v, 1]^{\top},
\end{equation}

This yields a \texttt{center\_3d} for each tracked mask. We then compute frame-wise gripper–object distances,
\begin{equation}
r_t(g,o)=\big\|\mathbf{p}^{(g)}_{\!t}-\mathbf{p}^{(o)}_{\!t}\big\|_2,
\end{equation}
for $g\in\{\text{gripper\_left,gripper\_right}\}$ and task-relevant objects $o$. For multi-view sequences, we build per-camera \texttt{relations\_3d} from the set of $r_t(g,o)$ values, and prefer the head camera if present; otherwise we select the camera with the most valid relations at a frame. We rely on \emph{temporal trends} in $r_t$ rather than absolute scale, avoiding the need for extrinsics.

\paragraph{LLM-Based Segmentation Conditioned on VLM Summaries.}
From the sampled frames (typically 4--8) and their structured 3D cues, we query a VLM (InternVL-3.5 8B~\cite{internvl3.5}) for a strict-JSON, chronologically ordered episode summary. We then attach this VLM-produced summary as a task descriptor to the LLM prompt. 

To enable the LLM (GPT-5 Thinking) to infer manipulation intent from raw 3D relations, we include a small set of few-shot exemplars that encode canonical temporal patterns, such as near-contact plateaus for precision phases and monotonic approach or retreat for coarse transit. These exemplars guide the model to interpret variations in the gripper–object distance $r_t$ as semantically meaningful interaction states.

We encode \emph{policy hints} to favor robust, human-like chunks. These hints act as soft priors to stabilize segmentation rather than fixed task-specific hyperparameters:
\begin{itemize}[]
\item \textbf{Intent criteria.} Sustained near-contact plateaus and low-variance micro-adjustments ⇒ precision;
long approach/retreat or persistent far separation ⇒ casual.
\item \textbf{Stability.}  Minimum segment length $L_{\min}{=}8$; merge same-label segments across gaps shorter than $G_{\min}{=}5$; require $\geq 3$ consecutive frames to switch labels (hysteresis); ignore micro-oscillations shorter than $L_{\mathrm{micro}}{=}6$.
\item \textbf{Parsimony.} Prefer $3$--$4$ segments unless strong evidence suggests otherwise.
\end{itemize}

Leveraging both the few-shot relational prior and the task description, the LLM infers temporal segment boundaries. 
The model receives (i) frame-wise \texttt{center\_3d} and \texttt{relations\_3d} for the episode and (ii) the VLM summary descriptor, and outputs non-overlapping index ranges labeled \texttt{precision} or \texttt{casual}. 
To ensure full coverage, we apply a deterministic post-processing step that fills small gaps by extending the nearest high-confidence segment consistent with the local $r_t$ trend. 
The final segmentation is represented as $\mathcal{S}=\{(s_i,e_i,y_i)\}_{i=1}^{M}$ with $y_i\in\{\texttt{precision},\texttt{casual}\}$.

For long demonstrations, we apply token-budgeted sampling. 
We select evenly spaced frames under a fixed character budget and compact the JSON representation through float rounding and whitespace removal, reducing prompt length by approximately 30--40\%.

\paragraph{Reproducible seed segmentation.}
While a human could segment the seed episode, our goal is a reproducible and scalable preprocessing mechanism.
Since casual–precision boundaries are inherently subjective and annotator-dependent, manual labeling does not provide a stable oracle at dataset scale.
ESPADA instead applies a consistent, geometry- and semantics-grounded policy to produce repeatable seed labels.

\subsection{Banded DTW Label Transfer from Episode-0}

For datasets where only episode~0 is labeled, we propagate its segment labels (\texttt{precision}/\texttt{casual}) to the remaining episodes by aligning trajectories using \emph{banded} Dynamic Time Warping (DTW). 
Because demonstrations of the same task typically differ in execution speed rather than overall motion structure, we restrict the warping path with a Sakoe--Chiba band so that only temporally nearby frames can be matched. 
This constraint preserves the global temporal ordering while allowing moderate speed variations across demonstrations.

\paragraph{Proprioceptive DTW Alignment.}
From each episode we build a per-frame feature vector using only proprioception and actions. Concretely, we concatenate z-scored features to form $\phi_t \in\mathbb{R}^{D}$:
\begin{multline}
\phi_t  \;=\; 
\big[\, a_t,\; \Delta a_t,\; v_t,\; \Delta v_t,\;
\|a_t\|,\;\|v_t\|,\;\|\Delta a_t\|, \\
\|\Delta q_t\|,\;\|\Delta v_t\|,\;
\angle(a_t,a_t{+}\Delta a_t),\;
\angle(v_t,v_t{+}\Delta v_t) \,\big].
\end{multline}
where $a_t$ are actions, $q_t$ are joint positions, $v_t$ are joint velocities if available (otherwise we use $\Delta q_t$ as a proxy), and $\angle(\cdot,\cdot)$ is the angle between successive vectors.

This feature representation was selected via ablation: among four variants ([$a_t$,$\Delta a_t$], +kinematics, +norms, +angles), the full representation in Eq.~(3) achieved the lowest mean absolute propagation error relative to the episode-0 ground truth. This representation combines kinematic cues for temporal motion evolution, norm-based cues for motion magnitude, and angular cues for directional consistency, yielding more discriminative DTW alignment across demonstrations with similar phase structure but different execution styles. 

Given episode~0 features $X_0\!\in\!\mathbb{R}^{T_0\times D}$ and target features $X_k\!\in\!\mathbb{R}^{T_k\times D}$ for episode $k$, we run DTW with a Sakoe--Chiba band of half-width
$b=\lfloor \rho\cdot\max(T_0,T_k)\rfloor$ with $\rho\!\in\![0.05,0.10]$ (default $\rho{=}0.08$). This yields an alignment path $\mathcal{P}\subset[1,T_0]\times[1,T_k]$.
We convert it into a monotone index map $m:\{1,\dots,T_0\}\!\to\!\{1,\dots,T_k\}$ by averaging all matched target indices per source frame and enforcing non-decreasingness.

\paragraph{Segment-wise Label Transfer and Refinement.}
For each episode-0 labeled segment $\mathcal{S}_0=\{(s_i,e_i,y_i)\}_{i=1}^{M}$ with label $y_i\in\{\texttt{precision},\texttt{casual}\}$, we obtain the target span $\mathcal{S}_k=\{(m(s_i),m(e_i),y_i)\}_{i=1}^{M}$ and snap both ends within a local window of $\pm W$ frames (default $W{=}12$) by minimizing the $\ell_2$ distance between short mean-pooled feature summaries.
Mapped segments are sorted and trimmed to remove overlaps while preserving order. If a path break occurs, we drop only the affected segment. Any uncovered frames default to \text{precise} when expanded to per-frame labels.
The banded DTW runtime is $\mathcal{O}(\max(T_0,T_k)\cdot b)$, i.e., near-linear in sequence length. With 50\,Hz episodes ($\sim$500–2k frames), transfers run quickly on CPU and require no proxy models.

\subsection{Segment-wise Downsampling and Dataset Compilation}
\label{sec:downsample}

Given the final segmentation $\mathcal{S}$, we construct an acceleration-aware dataset by applying RBD with a larger downsampling factor for casual spans and a smaller downsampling factor for precision spans.

\paragraph{Replicate-before-downsample (RBD).}
To maintain full state coverage under temporal compression, we adopt RBD~\cite{demospeedup}. For a segment $[s,e]$ and downsampling factor $N$, we create $N$ replicas with offsets $m\in\{0,\dots,N-1\}$ and retain frames $\{\,t\in[s,e] \mid (t-s)\bmod N=m\,\}$. Taking the union across $m$ recovers the original support, thereby preserving full state diversity in the downsampled dataset and preventing loss of observation coverage during model training.

\paragraph{Geometric Consistency for Chunked Policies.}\

Temporal acceleration alters the per-chunk spatial displacement, undermining the horizon $K$ that the policy has been optimized to perform best at.
To maintain geometric fidelity under accelerated demonstrations, we adopt the geometry-consistent downsampling scheme~\cite{demospeedup} and rescale the effective chunk horizon $K'$ so that its spatial displacement remains consistent with the original:
\begin{equation}
\sum_{k=0}^{K'-1}\big\|\Delta\mathbf{x}_{t+k}\big\|
\;\approx\;
\sum_{k=0}^{K-1}\big\|\Delta\mathbf{x}_{t+k}\big\|,
\end{equation}
where $\mathbf{x}_t$ denotes the end-effector pose. 
In practice, $K' \approx \tfrac{1}{2}K$ performs well and approximately satisfies Eq. (4) across tasks.

\paragraph{Gripper Event Precision Forcing.}
We apply gripper event precision forcing method to safeguard contact-rich phases from being over-accelerated. For each trajectory, we detect gripper movements by checking the change in the normalized gripper command $g_t$ and mark a frame as a candidate event if
$\lvert g_{t+4} - g_t \rvert \ge 0.03$.
All marked frames are then clustered along the temporal axis using DBSCAN~\cite{dbscan}. For each cluster, we take the minimum and maximum frame indices, pad them by two frames on both sides, and override the corresponding window to be precision on top of the base LLM segmentation results.








\section{Experiments}
We evaluate ESPADA along three dimensions: 
(1) whether it accelerates imitation-learning policies while maintaining task success, 
(2) whether it produces segmentation that better aligns with human interpretations of manipulation phases compared to entropy-based methods, and 
(3) how semantic cues such as 3D gripper–object distance and VLM-generated scene descriptions contribute to segmentation quality.

\subsection{Setup}
We evaluate our approach in both simulation and real-world settings using ACT and DP~\cite{act, dp} as the baseline policy architectures, and compare
our accelerated model against policies trained on the original dataset and those using the entropy-based acceleration method DemoSpeedup~\cite{demospeedup} under each architecture.

\textbf{Simulation.}
In Aloha simulation~\cite{alohaunleashed}, we evaluate two representative manipulation tasks—Transfer Cube and Insertion—each provided with 50 expert demonstrations at 50 Hz. Policies are trained from single head-camera observations. 
Experiments were conducted with precision/casual acceleration factors of (2x, 4x).
In BiGym~\cite{bigym}, we evaluate 7 long-horizon manipulation tasks that involve target reaching and articulated object interaction in home-like environments. 
Policies are trained with different numbers of demonstrations per task, while failed episodes are filtered out.

\textbf{Real-world.}
Experiments are conducted on the ROBOTIS AI-Worker~\cite{aiworker}, a dual-gripper humanoid robot equipped with two wrist-mounted cameras and a head-mounted camera. 
We evaluate five representative tasks—\textit{Sort}(bin sorting), \textit{Pen in Cup}(insertion), \textit{Kitchenware}(bowl and cup handling), \textit{Conveyor}(dynamic transfer) and \textit{Cloth}(deformable-object folding)—as shown in Fig.~\ref{fig:realworld-exp}, measuring both throughput and episode length across models. All policies follow the baseline-matched hyperparameters\cite{demospeedup} for both training and inference, and the accelerated segments use a chunk horizon of roughly half the original. 
DP exhibited limited robustness to large out-of-distribution deviations during preliminary experiments. To avoid conflating this effect with the impact of temporal acceleration, we reset the initial robot pose to lie within the training-time distribution for all tasks except \textit{Conveyor}.

\textbf{Metrics.} We evaluated whether time efficiency could be improved without compromising task success. We report the \textit{task completion success rate} and the \textit{average episode execution length}, where task failure is defined as the inability to proceed within 10 seconds in real-world experiments.

\begin{table*}[t]
\captionsetup{justification=raggedright,singlelinecheck=false}
\caption{BiGym Simulation Results. DemoSpeedup numbers are from~\cite{demospeedup}. Throughput is computed as success rate multiplied by normalized execution speed relative to the baseline policy.}
\label{tab:bigym_simulation_results}
\resizebox{\textwidth}{!}{%
\footnotesize
\begin{tabular}{l|cc|cc|cc|cc}
\hline
\textbf{Method} 
& \multicolumn{2}{c|}{\textbf{Sandwich Remove}} 
& \multicolumn{2}{c|}{\textbf{More Plate}} 
& \multicolumn{2}{c|}{\textbf{Load Cups}} 
& \multicolumn{2}{c}{\textbf{Put Cups}} \\
& success rate($\uparrow$) & episode len($\downarrow$) 
& success rate($\uparrow$) & episode len($\downarrow$)
& success rate($\uparrow$) & episode len($\downarrow$)
& success rate($\uparrow$) & episode len($\downarrow$) \\
\hline
ACT 
& 53\% & 368 
& \textbf{54\%} & 157 
& \textbf{61\%} & 319 
& 61\% & 288 \\

ACT-2x 
& 46\% & 193 
& 46\% & 119 
& 50\% & 195 
& 54\% & 141 \\

ACT+\textit{SAIL}
& 56\% & \textbf{142}
& 38\% & \textbf{86}
& 36\% &\textbf{113}
& 40\% & 152 \\

ACT+\textit{DemoSpeedup}
& 77\% & 156
& 53\% & 91
& 59\% & 176 
& \textbf{62\%} & \textbf{132} \\

\textbf{ACT+\textit{Ours}}
& \textbf{80\%} & 176 
& 24\% & 91
& 54\% & 173
& 60\% & 149 \\
\hline

DP 
& 52\% & 352 
& \textbf{52\%} & 170 
& 15\% & 419 
& 12\% & 386 \\

DP-2x 
& 51\% & 247 
& 41\% & 125 
& 11\% & 177 
& 7\% & 243 \\

DP+\textit{SAIL} 
& 16\% & \textbf{186}
& 44\% & 115
& 12\% & \textbf{139}
& 28\% & \textbf{180} \\

DP+\textit{DemoSpeedup} 
& \textbf{54\%} & 217 
& 49\% & 113 
& \textbf{38\%} & 171 
& 21\% & 205 \\

\textbf{DP+\textit{Ours}}
& 46\% & 200
& 40\% & \textbf{79} 
& 34\% & 162
& \textbf{38\%} & 218 \\
\hline
\end{tabular}
}

\vspace{0.3em}

\resizebox{\textwidth}{!}{%
\begin{tabular}{l|cc|cc|cc|ccc}
\hline
\textbf{Method} 
& \multicolumn{2}{c|}{\textbf{Saucepan to Hob}} 
& \multicolumn{2}{c|}{\textbf{Drawers Close}} 
& \multicolumn{2}{c|}{\textbf{Cupboard Open}}
& \multicolumn{3}{c}{\textbf{Averaged}} \\
& success rate($\uparrow$) & episode len($\downarrow$)
& success rate($\uparrow$) & episode len($\downarrow$)
& success rate($\uparrow$) & episode len($\downarrow$)
& success rate($\uparrow$) & speed-up($\uparrow$) & \textbf{throughput($\uparrow$)} \\
\hline

ACT 
& 86\% & 383 
& \textbf{100\%} & 119 
& \textbf{100\%} & 146 
& 74\% & 1.0$\times$ & 0.74 \\

ACT-2x 
& 81\% & 224 
& 87\% & 84 
& 96\% & 103 
& 66\% & 1.7$\times$ & 1.12\\

ACT+\textit{SAIL}
& 82\% & 159
& \textbf{100\%} & \textbf{39}
& 76\% & \textbf{62}
& 61\%& \textbf{2.4$\times$} & 1.44\\

ACT+\textit{DemoSpeedup}
& 92\% & 163 
& \textbf{100\%} & 63 
& \textbf{100\%} & 81
& \textbf{78\%} & 2.1$\times$ & 1.64\\

\textbf{ACT+\textit{Ours}}
& \textbf{94\%} & \textbf{148} 
& \textbf{100\%} & 56
& \textbf{100\%} & 81
& 73\% & 2.3$\times$ & \textbf{1.68}\\
\hline

DP 
& \textbf{79\%} & 324 
& \textbf{96\%} & 114 
& \textbf{100\%} & 181 
& 58\% & 1.0$\times$ & 0.58\\

DP-2x 
& 41\% & 242 
& 81\% & 65 
& 94\% & 161 
& 47\% & 1.5$\times$ & 0.71\\

DP+\textit{SAIL} 
& 76\% & 164
& 80\% & \textbf{39}
& 80\% & \textbf{81}
& 48\% & \textbf{2.2$\times$} & 1.03 \\

DP+\textit{DemoSpeedup} 
& \textbf{79\%} & 169 
& 89\% & 59
& \textbf{100\%} & 103
& \textbf{61\%} & 1.9$\times$ & 1.16\\

\textbf{DP+\textit{Ours}}
& 76\% & \textbf{148} 
& 88\% & 56
& \textbf{100\%} & 116 
& 60\% & 2.0$\times$ & \textbf{1.20}\\
\hline
\end{tabular}%
}

\end{table*}

\subsection{Simulation Results}
\label{sec:vb}
In Aloha simulation, as shown in Table~\ref{tab:Aloha Sim simulation_results}, naïve 2× acceleration reduces success rates, whereas our method maintains or even improves them while achieving up to 2.64× speedup over the original policy.
Compared with DemoSpeedup, our method achieves comparable overall performance on the \textit{Insertion} task while maintaining a similar level of acceleration, and attains the highest success rate on \textit{Transfer Cube} (ACT) despite being slightly less aggressive in shortening episode length.


\textbf{High Segmentation Quality Under Random Scenario.}
Random initialization of object positions in ALOHA increases entropy during the approach–grasp phase, causing entropy-based methods to mislabel interaction-critical spans as casual. To evaluate segmentation quality, we compare predicted segments with human-annotated ground-truth interaction phases.

\begin{figure}[t]  
    \centering
    \includegraphics[width=\columnwidth]{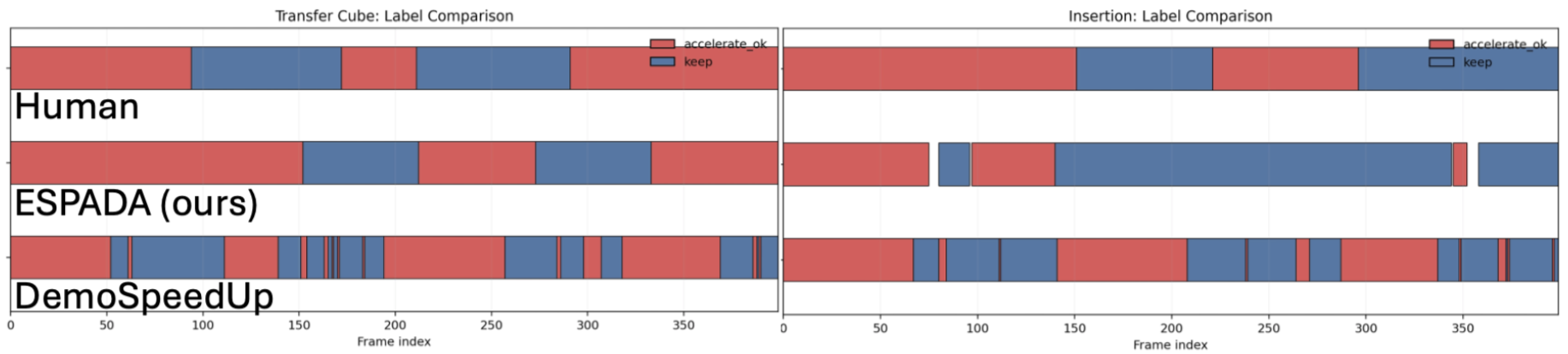}
    \caption{Segmentation comparison with human annotations. ESPADA closely matches human labels, while DemoSpeedup produces fragmented segments.
    }
    \label{fig:segmentation}
\end{figure}

As shown in Fig. \ref{fig:segmentation}, ESPADA produces boundaries that more closely align with human annotations, while DemoSpeedup often yields fragmented segments due to entropy fluctuations. Quantitatively, ESPADA achieves a substantially higher macro segment F1 IoU than DemoSpeedup (0.2609 vs.\ 0.1136 averaged across tasks). Similar improvements are observed on both tasks: 0.2848 vs.\ 0.1295 for \textit{Transfer Cube}, and 0.2369 vs.\ 0.0977 for \textit{Insertion}.

This improved segmentation also translates to better early-phase interaction detection: in the initial contact-detection subtask, ESPADA achieves 91\% success compared to 87\% for the entropy-based baseline.

\begin{table}[t]

\captionsetup{justification=raggedright,singlelinecheck=false}
\caption{Aloha-Sim Simulation Results.}
\label{tab:Aloha Sim simulation_results}

\resizebox{\columnwidth}{!}{%
\begin{tabular}{l|cc|cc|c}
\hline
\textbf{Method} 
& \multicolumn{2}{c|}{\textbf{Insertion}} 
& \multicolumn{2}{c|}{\textbf{Transfer Cube}} 
& \textbf{throughput}($\uparrow$) \\
& success rate($\uparrow$) & episode len($\downarrow$) 
& success rate($\uparrow$) & episode len($\downarrow$) \\
\hline

ACT 
& 21\% & 452 
& \textbf{72\%} & 291 
& 0.47 \\

ACT-2x 
& 13\% & 238 
& 70\% & 162 
& 0.75 \\

ACT+\textit{DemoSpeedup}$_\text{(repro)}$ 
& \textbf{28\%} & \textbf{166} 
& 66\% & \textbf{127} 
& \textbf{1.13} \\

\textbf{ACT+\textit{Ours}}
& \textbf{28\%} & 171 
& \textbf{72\%} & 141 
& 1.12 \\

\noalign{\smallskip}\hline\noalign{\smallskip}

DP 
& 16\% & 431 
& \textbf{66\%} & 281 
& 0.41 \\

DP-2x 
& 12\% & 245 
& 61\% & 146 
& 0.69 \\

DP+\textit{DemoSpeedup}$_\text{(repro)}$  
& \textbf{26\%} & \textbf{173} 
& 64\% & \textbf{137} 
& \textbf{0.98} \\

\textbf{DP+\textit{Ours}}
& \textbf{26\%} & 193 
& 58\% & 121 
& 0.97 \\

\hline
\end{tabular}%
}

\end{table}

\textbf{Long-Horizon Speedup and Sensitivity to Unstable Visual Scenes.}
In BiGym, our method achieves higher success rates than the simple 2× baseline (ACT: 66\%→73\%, DP: 47\%→60\%) while maintaining performance comparable to the original 1× policy and providing up to 2.3× acceleration. 
When measured using throughput (success rate × normalized execution speed), our approach achieves the highest values among all methods (ACT: \textbf{1.68}, DP: \textbf{1.20}), indicating the best trade-off between task success and execution efficiency.
Interestingly, faster execution often improved success by reducing compounding errors and preventing drift into OOD states. 
We also include SAIL as an additional baseline; although it achieves strong compression in some tasks, its clustering-based segmentation yields lower success rates overall, highlighting the limitations of heuristic phase proxies in diverse manipulation scenarios.
Finally, some failures appear to stem from unstable visual observations—often when the robot briefly moves outside the object scene—which can degrade gripper–object recognition and VLM grounding. We leave improving robustness through more stable viewpoints and richer multimodal signals (e.g., joint states or haptics) to future work.

\begin{table*}[h!]
\caption{Real-world Results.}
\label{tab:realworld_results}
\resizebox{\textwidth}{!}{%
\Large
\begin{tabular}{l|cc|cc|cc|cc|cc|ccc}
\hline
\textbf{Method} & \multicolumn{2}{c|}{\textbf{Pen in Cup}} & \multicolumn{2}{c|}{\textbf{Sort}} & \multicolumn{2}{c|}{\textbf{Kitchenware}} & \multicolumn{2}{c|}{\textbf{Conveyor}} & \multicolumn{2}{c|}{\textbf{Cloth}} & \multicolumn{2}{c}{\textbf{Averaged}} \\
& success rate($\uparrow$) & episode len($\downarrow$) 
& success rate($\uparrow$) & episode len($\downarrow$) 
& success rate($\uparrow$) & episode len($\downarrow$) 
& success rate($\uparrow$) & episode len($\downarrow$)
& success rate($\uparrow$) & episode len($\downarrow$) 
& success rate($\uparrow$) & speed-up($\uparrow$) & \textbf{throughput($\uparrow$)} \\
\hline
\noalign{\smallskip}
ACT 
& \textbf{29/30} & 18.67 
& 27/30 & 37.52 
& 8/20 & 38.68 
& 13/20 & 9.89 
& \textbf{9/10} & 77
& 76.3\% & 1.0x & 0.76 \\

ACT+\textit{DemoSpeedup} ~ 1/3 
& \textbf{29/30} & 15.52 
& \textbf{29/30} & 29.29 
& 10/20 & 25.62 
& 4/20 & 7.55 
& 0/10 & x
& 52.7\% & 1.34x & 0.71\\

ACT+\textit{DemoSpeedup} ~ 2/4 
& 27/30 & \textbf{5.36} 
& 24/30 & \textbf{13.23} 
& 1/20 & \textbf{17.39} 
& 1/20 & \textbf{4.49} 
& 0/10 & x
& 36.0\% & \textbf{2.59x} & 0.93 \\

ACT+\textit{Ours} ~ 1/3 
& \textbf{29/30} & 15.32 
& \textbf{29/30} & 29.32 
& 10/20 & 22.76 
& \textbf{19/20} & 7.36 
& 7/10 & 63.99
& 81.7\% & 1.35x & 1.10 \\

ACT+\textit{Ours} ~ 2/4 
& \textbf{29/30} & 6.57 
& 28/30 & 15.56 
& \textbf{16/20} & 20.72 
& 18/20 & 4.51 
& \textbf{9/10} & \textbf{36.88}
& \textbf{90.0\%} & 2.26x & \textbf{2.03}\\

\noalign{\smallskip}
\hline
\noalign{\smallskip}

DP 
& 11/15 & 21.55 
& 10/15 & 48.29 
& 0/15 & x 
& 0/20 & x 
& 5/10 & 35.14
& 38.0\% & 1.0x & 0.38 \\

DP+\textit{DemoSpeedup} ~ 2/4 
& \textbf{15/15} & 8.66 
& 13/15 & 23.58 
& \textbf{10/15} & 27.50 
& 13/20 & \textbf{6.15} 
& 3/10 & \textbf{11.57}
& 69.7\% & 2.52x & 1.76\\

DP+\textit{Ours} ~ 2/4 
& \textbf{15/15} & \textbf{5.83} 
& \textbf{15/15} & \textbf{21.54} 
& \textbf{10/15} & \textbf{23.15} 
& \textbf{15/20} & 7.41 
& \textbf{7/10} & 19.21
& \textbf{82.3\%} & \textbf{2.59x} & \textbf{2.13} \\
\noalign{\smallskip}
\hline
\end{tabular}%
}
\end{table*}

\begin{figure}[t]  
    \centering
    \includegraphics[width=0.8\columnwidth]{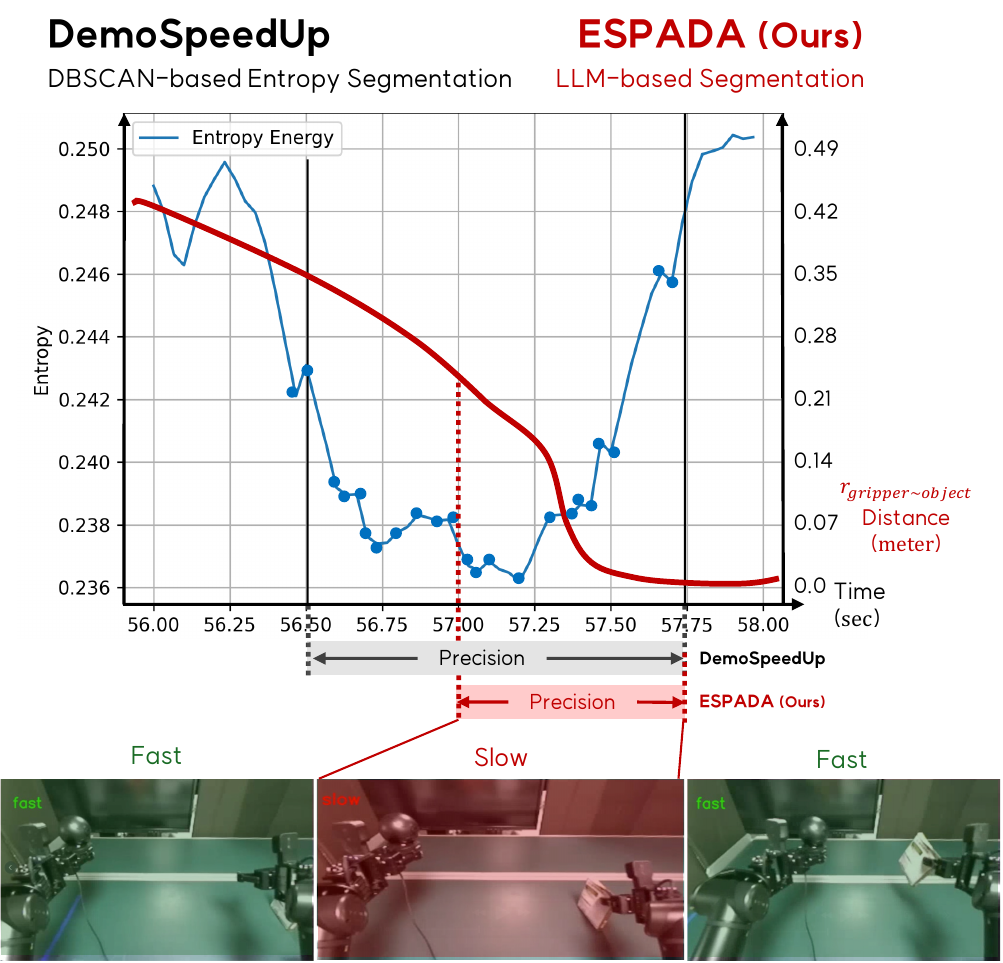}
    \caption{\textbf{Precision-phase estimation in the conveyor scenario based on low entropy (DemoSpeedup, black) versus semantics (Ours, red).}
DemoSpeedup often enters the precision phase too early due to low action entropy in repetitive approach motions, leading to premature slowdown. Our method delays precision until the brief contact alignment with the moving object, enabling reliable grasping.
    }
    \label{fig:espada-vs-dsu}
\end{figure}

\subsection{Real-world Results}

As shown in Table~\ref{tab:realworld_results}, we evaluate real-world performance under the 2$\times$/4$\times$ acceleration setting, where precision segments are accelerated by $2\times$ and casual segments by $4\times$. Under this setting, ESPADA achieves the highest overall success rates while maintaining substantial execution speedup across all tasks.

For ACT, ESPADA achieves \textbf{90.0\%} success at \textbf{2.26$\times$} speedup, whereas DemoSpeedup drops to 36.0\% despite achieving a marginally higher nominal acceleration by aggressively classifying many spans as casual. This over-acceleration is particularly evident in the \textit{Conveyor} task, where DemoSpeedup collapses to 1/20 success while ESPADA maintains 18–19/20 successful executions.

A similar trend holds for DP. ESPADA achieves the highest success rate (82.3\%) together with the largest speedup (2.59$\times$), outperforming DP+DemoSpeedup (69.7\%).

Notably, ESPADA also remains robust in the deformable-object \textit{Cloth} task, maintaining stable performance (9/10 successes under ACT and 7/10 under DP). Here, we use a lightweight task-adapted relation based on the nearest point on the cloth boundary, suggesting that semantic phase segmentation generalizes beyond rigid-object manipulation.

\textbf{Casual-exploiting Segmentation.}
By combining temporal trends in the gripper–object distance with VLM-generated scene descriptions, ESPADA reliably identifies genuine precision phases (e.g., near-contact adjustments) while aggressively compressing truly casual spans. In contrast, entropy-based segmentation treats low action entropy as a proxy for precision. This assumption fails in repetitive motions, where entropy may remain low even without fine control, causing DemoSpeedup to overestimate precision-critical spans. As shown in Fig.~\ref{fig:espada-vs-dsu} for the \textit{Conveyor} task, this misclassification marks large portions of repetitive scooping as non-accelerable, limiting potential speed gains and obscuring casual segments that ESPADA correctly recovers.

The same trend appears quantitatively in the 1/3-acceleration setting. ESPADA achieves shorter episode lengths not by compressing true precision phases, but by avoiding DemoSpeedup’s overextension of low-entropy repetitive segments. Across tasks, ESPADA preserves success-critical precision phases while more accurately identifying accelerable casual spans.

\textbf{Precision-preserving Segmentation.}
In \textit{Kitchenware} (ACT, 2×/4×), ESPADA achieves \textbf{16/20} successes versus \textbf{1/20} for DemoSpeedup, showing that ESPADA reliably preserves precision-critical phases even under strong acceleration. 

In contact-rich manipulation, near-contact spans must not be down-sampled; ESPADA preserves this precision-critical interaction, whereas DemoSpeedup down-samples the delicate cup-grasp phase too aggressively, leading to only 1/20 success. The gripper–object-distance trend fed into the LLM allows it to infer phase intent (approach → align → close), while conservatively gating gripper events, thereby retaining precision spans and avoiding compression of delicate motions.

A similar behavior appears in the deformable-object \textit{Cloth} task. Although the gripper remains in continuous contact with the cloth, ESPADA still identifies the interaction phases and maintains stable performance (9/10 successes under ACT and 7/10 under DP), indicating that the segmentation remains effective even for dexterous deformable manipulation.

\textbf{Robustness in Dynamic Scenario.}
The \textit{Conveyor} task highlights limitations of entropy-based acceleration in dynamic scenes. When the target object (curry) first enters the camera view, the robot must remain still and wait for the correct pickup configuration. However, action entropy is naturally high during this transient phase, causing DemoSpeedup to misclassify it as casual and trigger arm descent prematurely. This often leads to unrecoverable joint configurations under ACT’s joint-state-conditioned action chunking.

In contrast, ESPADA identifies this waiting phase as precision using semantic cues from VLM descriptions together with the temporal trend of the gripper–object distance. The robot therefore waits until the object reaches the correct pickup zone, enabling stable execution even under irregular conveyor timing.

\subsection{Ablations}

\paragraph{Semantic cue ablation.}
We ablate the effect of gripper–object distance \( r \) and the VLM scene description using four variants: w/o \( r \), w/o description, w/o both, and our full model (Table~\ref{tab:ablation_combined}). We report IoU and the predicted number of segments with respect to the ground-truth segmentation.

\begin{table}[h!]
    \centering
    \caption{Ablation on IoU and number of segments.}
    \label{tab:ablation_combined}
    \begin{tabular}{lcccc}
        \hline
        \multirow{2}{*}{\textbf{Method}} 
            & \multicolumn{2}{c}{\textbf{Insertion}} 
            & \multicolumn{2}{c}{\textbf{Transfer Cube}} \\
        \cmidrule(lr){2-3} \cmidrule(lr){4-5}
            & IoU & \#Seg. & IoU & \#Seg. \\
        \hline
        w/o \( r \)        & 0.0224 & 3/3 & 0.2791 & 1/2 \\
        w/o Desc.          & 0.1024 & 3/3 & 0.0584 & 3/2 \\
        w/o \( r \), Desc. & 0.1111 & 3/3 & 0.0693 & 3/2 \\
        Ours               & \textbf{0.5166} & 3/3 & \textbf{0.3064} & 2/2 \\
        \hline
    \end{tabular}
\end{table}

For \textbf{Insertion}, removing \( r \) collapses IoU (0.5166 → 0.0224), indicating that $r$ is essential for alignment-sensitive interactions. For \textbf{Transfer Cube}, dropping the description sharply reduces IoU (0.3064 → 0.0584), suggesting that textual cues help disambiguate phases with similar geometry. All variants recover the correct number of segments, but only our full model achieves tight temporal alignment. Overall, this ablation confirms that both spatial relations and semantic descriptions are necessary for high-fidelity segmentation.

\paragraph{Gripper-event ablation.}
We also ablate the gripper-event signal used for precision-phase forcing. Removing this signal leads to moderate performance degradation (e.g., ACT Transfer Cube 72\%→62\%, DP Insertion 26\%→10\%), indicating that gripper events provide helpful semantic cues for interaction phases but are not required for ESPADA to function.

\paragraph{Aggressive acceleration study.}
We further test a more aggressive setting in ALOHA, accelerating precision segments by $4\times$ and casual segments by $6\times$. Under this setting, ESPADA maintains higher success rates than DemoSpeedup while achieving similar episode lengths (Insertion: 36\% vs.\ 32\%, Transfer Cube: 38\% vs.\ 26\%). When measured using throughput (success rate × normalized speed), ESPADA also achieves higher values (1.52 vs.\ 1.38 and 1.54 vs.\ 1.05). This indicates that ESPADA’s segmentation enables stronger temporal compression without degrading policy performance.

\section{Conclusion}
We presented ESPADA (Execution Speedup via Spatially Aware Demonstration Data Downsampling), a semantic segmentation framework that accelerates demonstrations without requiring additional data, hardware, or policy retraining. By exploiting scene semantics and gripper–object spatial relations, ESPADA distinguishes accelerable from precision-critical segments, producing motion-aligned chunks and reducing temporal redundancy via RBD with geometric consistency. Integrated with ACT and DP, ESPADA achieves natural motion chunking, preserves task success, and generalizes across both simulation and real hardware.

\textbf{Limitations.} ESPADA still faces challenges: inaccurate masks or object tracking may distort spatial relations, monocular depth estimation introduces noise, and further validation is needed for large-scale deployment. DTW-based label transfer assumes similar subtask ordering across demonstrations; substantially different strategies may require additional seed segmentations. Addressing these issues will be crucial for advancing ESPADA as a reliable and general framework for safe and efficient policy acceleration.

\bibliographystyle{IEEEtran}
\bibliography{references}

\end{document}